\documentclass[17pt]{article}

\usepackage[utf8]{inputenc} 
\usepackage[T1]{fontenc}    
\usepackage{url}            
\usepackage{booktabs}       
\usepackage{amsfonts}       
\usepackage{nicefrac}       
\usepackage{microtype}      
\usepackage{xcolor}         
\usepackage{amsmath}
\usepackage{amssymb}
\usepackage{algorithm}
\usepackage{geometry}
\usepackage{algorithmic}
\usepackage{graphicx}
\usepackage{subfig}
\usepackage{xcolor}
\usepackage{pagecolor}
\usepackage{lipsum}  
\PassOptionsToPackage{square,numbers}{natbib}
\date{}
\title{DSDF: An approach to handle stochastic agents in collaborative multi-agent reinforcement learning}

\author{%
  Satheesh K. Perepu and Kaushik Dey\\
  Ericsson Research (Artificial Intelligence)\\
  Chennai, Tamil Nadi\\
  India 600096 \\
  \texttt{perepu.satheesh.kumar@ericsson.com, deykaushik@ericsson.com} \\
}

\begin{document}

\maketitle

\begin{abstract}
  Multi-Agent reinforcement learning has received lot of attention in recent years and have applications in many different areas. Existing methods involving Centralized Training and Decentralized execution, attempts to train the agents towards learning a pattern of coordinated actions to arrive at optimal joint policy. However if some agents are stochastic to varying degrees of stochasticity, the above methods often fail to converge and provides poor coordination among agents. In this paper we show how this stochasticity of agents, which could be a result of malfunction  or aging of robots, can add to the uncertainty in coordination and there contribute to unsatisfactory global coordination. In this case, the deterministic agents have to understand the behavior and limitations of the  stochastic agents while  arriving at optimal joint policy. Our solution, DSDF which tunes the discounted factor for the agents according to uncertainty and use the values to update the utility networks of individual agents. DSDF also helps in imparting an extent of reliability in coordination thereby granting stochastic agents tasks which are immediate and of shorter trajectory with deterministic ones taking the tasks which involve longer planning. Such an method enables joint co-ordinations of agents some of which may be partially performing and thereby can reduce or delay the investment of agent/robot replacement in many circumstances. Results on benchmark environment for different scenarios shows the efficacy of the proposed approach when compared with existing approaches.
\end{abstract}
\section{Introduction}
Multi-agent reinforcement learning (MARL) has been applied to wide variety of application which involve participation of collaborative agents such as Traffic management \cite{paper:MARL_App_Traffic}, power distribution \cite{paper:MARL_App_Power}, fleet management \cite{paper:MARL_App_Fleet} etc. There are different ways in to achieve this collaboration. Some set of algorithms focus on learning a centralized policies \cite{paper:Deep_MARL_Cent,paper:Deep_MAR_Cent2} while some on decentralized policies \cite{paper:MARL_Dec}. To improve the performance of the decentralized policies, some works assumed centralized training while learning these policies \cite{paper:IQL,paper:QMIX,paper:maven}.

The first multi-agent methods were based on updation of Q-values arranged in form of table \cite{paper:MARL_Tabel}.  This approach is not scalable with increase in number of states and/or actions as one need to store all these values in memory. With the advent of deep networks, people started modelling value function using deep Q networks \cite{paper:deeprl,paper:deeprl2,paper:MARL_Deep1,paper:MARL_Deep2,paper:MARL_Deep3,paper:MARL_Deep4}. In literature, people call these approaches under the name of deep Q learning. 

Deep Reinforcement learning has been effective to deal with challenges of Multi-Agent coordination. The simplest technique was to forego a centralized coordination and let the agents learn independent value functions known as independent Q-learning (IQL) \cite{paper:IQL}. However the same does not handle non-stationarity introduced by actions executed by other agents. Although there exists some variants of IQl like \cite{paper:IQL_Variant}, they miss some vital information on how other agents are performing. With introduction of deep Q-learning in Multi-Agent problems \cite{paper:MARL_Deep10} complex environment representations were possible but the problems of non-stationarity still persisted. Subsequently, with introduction of centralized experience replay \cite{paper:Deep_MARL_Cent,paper:Deep_MAR_Cent2} the problem of coordination started to be addressed using centralized approach. Further, some recent work involving COMA \cite{paper:coma} which uses a centralized critic and actors with counterfactuals to address non stationarity and credit assignment has proved to effectively address Multi-agent coordination. However this requires on-policy learning and thus more number of samples to learn optimal joint policy. Also, this method is not easily scalable with number of agents. 

Sunehag et. al. \cite{paper:VDN} proposes a value decomposition method (VDN) which updates the Q-networks of individual agents based on sum of all the agents value function using centralized training and decentralized execution. However the same may not consider the extra state information of the environment and also it cannot be applied to all the general MARL problems and particularly where joint Q function is not a linear function of individual Q functions. To address this problem, Rashid et. al. \cite{paper:QMIX} proposed a QMIX method which lies between extremes of VDN and COMA. The proposed approach uses a mixing network which mixes the individual agents value function through a mixing network which is then used to obtain  $Q_{tot}$. Further the agents' value function are trained based on the $Q_{tot}$ and the mixing network is trained by conditioning the same on the state of the environment. In this way we can add the state of the environment while training of the individual agents and also the agents are trained based on other agents performance. However, the issue with this approach is that the value function of individual agents should monotonically increase w.r.t the $Q_{tot}$. To handle the stricter non-monotonic assumption, Mahajan et. al. \cite{paper:maven} proposed a method known as MAVEN to train decentralized policies for agents condition their behavior on the shared latent variable controlled by a hierarchical policy. 

All these above methods generally assume the agents behave exactly in the way as policy instructed it. However in many cases, the agents can behave in entirely stochastic way i.e. they execute the actions different that of the actions given by the policy. The degree of stochasticity can be different from different agents and over the time it can become constant. This is a common phenomena in industries where we learn policies using agents which when they get old and undergo wear and tear may not always be able to follow the strict demands of the policy. To explain it better let us consider the following example. 

Let us assume there are robots in a warehouse performing some task. At the start of the experiment we have all the robots performing up to expectations and assume we train the agents to learn a collaborative policy to perform a task. However over the time robots can go under wear and tear and need not perform according to expectations. For example, at the start of the experiment applying an action ``forward" to motors move the robot forward with a speed of 16 KM/HR. However, over the time motors can undergo wear and tear and motors may not perform well. Now, with the same action of ``forward" results in moving some robots forward with a speeds varying between 8 KM/HR to 16 KM/HR while some may still perform per expectations. As one can see the joint policy learned for the previous case will no longer perform well for this case. It should be noted that only some of the agents behave in stochastic way as it will depend on the factors such as extent of usage, defects during manufacturing etc. One solution is to replace those stochastic agents but it is not recommended as it involves some cost with replacement.

Hence we need to retrain the policy from scratch when some of the agents are behaving in this stochastic way.  Also, the changes in the agent won't be drastic and stays at the same degree of stochasticity for some period of time. Hence it can be concluded that the behavior of agents will be same for both training and execution phase and the policy learnt during training will work for execution phase also for a reasonable amount of time. 

Intuitively, the success of coordination will depend on the deterministic (and good) agents understanding the limitations of stochastic agents, their behavior and tune the respective decentralized execution policies such that reward of overall joint policy is maximized. In this work, we achieve this coordination by using different discounted factors for different agents so that the respective dependencies of the value function on the future values is effected appropriately. For highly stochastic agents, we will use lower discounted factor so that their value function will depend on short-term actions and not so much on the long-term actions. For deterministic agents we will choose higher discounted factor to make them more depend on future values and let them plan effectively towards long term strategies. 

In this paper we came with two methods to obtain the discounted factor for all the agents. One is the iterative penalization method which will penalize the discounted factor for the agent for every stochastic action taken. Another is to use a Deep Stochastic Discounted Factor (DSDF) method which will predict the discounted factor based on local observations and state of the environment. More details on both the methods are proposed in Section \ref{sec:Proposed}. For the sake of comparison, we compared the results with QMIX \cite{paper:QMIX} and IQL \cite{paper:IQL} and corresponding plots are analyzed in Section \ref{sec:Proposed}.
\section{Background}
In this work, we assume a fully cooperative multi-agent task which is described with a decentralized partially observable Markov decision process (Dec-POMDP) which is defined with a tuple $G = <S,\mathbf{U},P,r,Z,O,N,\gamma>$ where $s \in S$ describes the state of the environment \cite{paper:DEC-MDP}. At each step in time, each agent $i$ out of these $N$ agents will take an action $a_i$ and for all the agents the join action is represented by $U$. Due to the applied actions, the system will transit to  $P(s'|s,\mathbf{U}): \mathbf{S} \times \mathbf{U} \times \mathbf{S} \longrightarrow [0,1] $. All the agents share common reward function $r(s,\mathbf{U}: \mathbf{S}\times \mathbf{U} \in \mathcal{R}$ and $\gamma:N \times 1 \longrightarrow [0,1]$ is the discounted factor chosen for N agents. 

Here we consider the environment is partially observable in which agent draws individual observation $z \in \mathbf{Z}$ according to a observation space $O(s,a): \mathbf{S}\times \mathbf{A} \longrightarrow \mathbf{Z}$. Each agent can have their own observation history $\tau_i \in \tau: (\mathbf{Z}\times \mathbf{U}$ which influences the underlying stochastic policy $\pi^i(u^i|\tau^i)$. The joint policy $\pi$ has a join action-value function $Q^\pi(s_t,u_t) = E_{s_{t+1:\infty},u_{t:\infty}}[R_t|S_t,u_t]$, where $R_t=\displaystyle \sum_{i=0}^\infty \gamma^i r_{t+i}$.
\section{Proposed method}\label{sec:Proposed}
In this section, we describe the proposed approach to handle the stochastic agents in MARL approach. As described earlier, the proposed approach results in learning the most optimal joint policy of all agents even when some agents are stochastic. An important assumption here is ``All the agents are deterministic while starting the experiment. However, after sometime the agents may become stochastic and we need to retrain the policy from scratch so that we address these stochastic agents. During the execution phase also we assume the same agents are stochastic with same degree of stochasticity. This is required to prevent the usage of continuous learning during execution."

The proposed approach is explained with QMIX \cite{paper:QMIX} as collaborative mechanism between agents since it is considered as one of state of art collaborative mechanism. Although it can be extended to any other collaborative mechanisms which satisfies centralized training and decentralized execution. 

In QMIX we compute the utility function for agent $i$ out of all $N$ agents as $Q_i(\tau_i,a_i)$, where $\tau_i$ is the history of observation space and action for agent $i$. The agent utility values $Q_i(\tau_i,a_i)$ are obtained by sending the observation spaces $(o_i,a_i)$ through a network with parameters $\theta^i$, where $i$ is the agent index out of $N$ agents. The obtained utility values $Q_i(\mathbf{o}_i,a_i)\;\;\forall \;\; i = 1,\cdots,N$ are mixed using a mixing network to obtain combined mixing value $Q_\text{tot}(\tau,\mathbf{U})$. Now, each agent $i$ network $\theta^i$ is updated by total value function $Q_\text{tot}$ instead of local utility function $Q_i(\mathbf{o}_i,a_i)$. The advantage here is that each agent network will have information on reward obtained and also indirectly other agent performance. In this way, we can incorporate other agents information without actually collecting them and able to arrive at joint policy. 

In normal circumstances where all the agents are deterministic the agents together learn a joint policy since all the agents will know how other agents will work. However in the case where some agents are stochastic, the deterministic agents have to adjust to other agents stochastic behavior and tune their own behavior to arrive at good joint policy. This can be achieved by tuning the value function of stochastic agents to depend only on the current values and not on the future values. This can be achieved through the tuning the value of discounted factor for each agent.  

However, in existing approach it assumes all agents are deterministic and we use a single discounted factor. However when some of the agents are stochastic using a single discounted factor can often results in poor update of network. For example, if the agent is highly stochastic, then it will take random actions with high probability which can be different from the action chosen by the policy. Now, if we use higher discounted factor here to update the network, it will result in poor update as the future actions can be totally different from planned actions (even in the case of exploration) with high probability and the same would upset the planned coordination of all the agents. Now, ideally we need to use smaller value of discounted factor for the higher stochastic agents and vice-versa to restrict the value function of those stochastic agents to short-term value rather than the future long-term value. 

Hence, in this work we propose to compute the discounted factor for each agent based on the current observation and also the global state of the environment. In this work we propose two methods to compute the optimal $\gamma_i$ value for agent $i$. (i) Iterative penalization method where we penalize the discounted factor of each agent for every stochastic action taken and (ii) Fully connected network to compute the optimal discounted factors for each agent. For this we train a fully connected neural network which will output the discounted factor values for each agent $i$. In both the above cases, the computed discounted factor is inputted to the individual utility networks and use the computed discounted factor to compute the $y$, where $y = r + Q_{tot}(s',\mathbf{u},\theta^-,\gamma)$ $\theta^-$ is the target network. The previous equation can be rewritten as  $y = r + g(s',\gamma_iQ_i(o_i,a_i),\mathbf{u},\theta_{tot})$, where $\theta_{tot}$ is the parameters of mixing network. Now, we update both the utility networks, mixing network (hypernetwork to be exact) using the different discounted factors for different agents. In this way, we can handle stochastic agents. 

\subsection{Proposed methods to calculate optimal $\gamma$ for all the agents}
In this work, we propose two methods to compute the optimal $\gamma$ for all the agents. Below we explain them in detail.

\subsubsection{Iterative penalization Method}\label{sec:simple_method}

In this method, we assume the discounted factor is $1$ for all the agents during the starting of the experiment i.e. during retraining from scratch. If the action executed by the agent $i$ is different from that of action given by the policy we will penalize the discounted factor for the agent $i$ by a factor $P$. At every one time step,  with every mismatch we will decrease the discounted factor with the factor $P$. Now, we will use the latest discounted factor at the time step to compute the utility function. Finally, we use the regular QMIX approach to update the utility networks and mixing networks.

The penalization factor $P$ should decrease with time steps just like we do in exploration factor in $\epsilon$-greedy method. The choice of optimal value for $P$ can be itself posed as optimization problem which is out of scope of the paper. However, the method proposed in Section \ref{sec:Neural} does not require any hyperparameter selection.  

\subsubsection{Deep Stochastic Discounted Factor (DSDF) Method}\label{sec:Neural}
\begin{figure}
    \centering
    \includegraphics[scale=0.35]{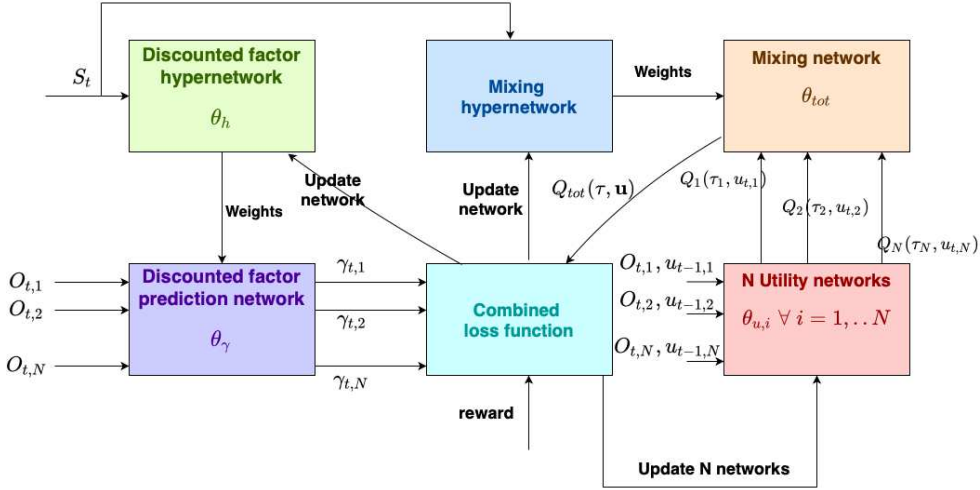}
    \caption{Proposed approach to compute discounted factor}
    \label{fig:Discounted_Factor}
\end{figure}

In this case, we proposed a method to compute the optimal $\gamma_i,\;\; i = 1,\cdots,N$ using a trained network. The proposed method is shown the Figure \ref{fig:Discounted_Factor}. 

The idea is to utilize the agent local observations $o_{t,i}$ and global state $s_t$ at time instant $t$ of the environment to compute the discounted factor for individual agents. The reason behind is explained below:

Since we assume the underlying collaborative mechanism works for deterministic agents, the only issue behind poor results is due to the stochasticity of agents. Each agent have their perspective of the global state in form of local observations. Hence local observations will have an idea on the stochasticity of the agent and we need to estimate the discounted factor depending on the local observations and global state of the environment. 

Returning to the main discussion, we use the local observations of agents to estimate the discounted factors using a fully connected network. However, we require additional information about the global state to the process. Since the local observations and global state are in different scale, we cannot combine together in same network. Also, since we require the discounted factor values between $0$ to $1$ from the network, we cannot train this network directly. Hence we use the concept of hypernetwork described in \cite{paper:hypernetworks}. 

The local observations of all $N$ agents are sent to the network $\theta_\gamma$ which will compute the discounted factor values $\gamma_i,\;\;i = 1,\cdots,N$. We utilize the hypernetwork $\theta_h$ to arrive at the network weights $\theta_\gamma$. The training process to update the $\theta_h$ is explained below. 

As one can see the $\theta_h$, $\theta_u$ (utility networks) and $\theta_m$ are interlinked with each other. The general loss function of QMIX is
\begin{align}
    \mathcal{L}(\theta) = \displaystyle \sum_{t=1}^{B}\left(y^{t}-Q_{tot}(\tau,\mathbf{u},s:\theta\right)
    \label{eq:DQN}
\end{align}
where $\theta$ is the set of parameters of $N$ utility agents and mixing hypernetwork. Now, if we expand $y^{t}$
\begin{align}
    y^{t} & = r + \gamma \;\; \underset{u'}{\max}\;\; Q_{tot}(\tau',\mathbf{u}',s':\theta^{-}) 
\end{align}
Now, instead of using single $\gamma$ value we will take the $\gamma$ inside the equation and we use our predicted network to compute the discounted factor. 
\begin{align}
    y^{t} & = r + \underset{u'}{\max} \;\;g(\mathbf{u}',s',\gamma_iQ_{u,i},\gamma_{tot}) 
\end{align}
Here $g(.)$ is the mixing network architecture which is parametrized by $\theta_{tot}$, $Q_{u,i}$ is the individual utility function of agent $i$. Now, we will replace the $\gamma_i$ with output of the network $\theta_\gamma$. The replaced equation is 
\begin{align}
    y^{t} & = r + \underset{u'}{\max}\;\; g(\mathbf{u}',s',f_\gamma(o_1^t,\cdots,o_N^t,\theta_\gamma)Q_{u,i},\theta_{tot}) \nonumber\\
    & = r + \underset{u'}{\max}\;\; g(\mathbf{u}',s',f_\gamma(o_1^t,\cdots,o_N^t,f_h(\theta_h,s'))Q_{u,i},\theta_{tot})
    \label{eq:y_tot}
\end{align}
where $f_\gamma$ is the discounted factor hyper network which is parametrized by $\theta_\gamma$ and $f_h$ is the hypernetwork function which is parametrized by $\theta_h$. 

Replacing the value of $ y_i^{tot}$ from \eqref{eq:y_tot} with that of \eqref{eq:DQN} we obtain the loss function as 
\begin{align}
    \mathcal{L}(\theta, \theta_h) = \displaystyle \sum_{t=1}^{B}\left(r^t + \underset{u'}{\max} g(\mathbf{u}',s',f_\gamma(o_1^t,\cdots,o_N^t,f_h(\theta_h,s'))Q_{u,i},\theta)-Q_{tot}(\tau,\mathbf{u},s:\theta\right)
    \label{eq:DQN_New}
\end{align}
There are two unknown parameters in the above equation (i) $\theta$, (ii) $\theta_h$. Since the parameters are interdependent on each other i.e. $\theta_h$ on $\theta$ and vice-versa, we need to solve them in an iterative fashion. For every $B$ samples, first we will update the hyper network $\theta_h$ for fixed $\theta$ and then 
update $\theta$ for the computed $\theta_h$. So at each step we update the both $\theta_h$ and $\theta$ iteratively.

A important point to note here is that $\theta$ needs to be updated for every batch of samples as each batch will have new information on the environment. However, since the degree of stochasticity of the agents are assumed to be constant the $\theta_\gamma$ network will converge after some iterations and hence we don't want to update the $\theta_\gamma$ after some batches. Hence, it may be noted that we are adding only one training step to the existing mechanism for some batches. It can be inferred that we are adding only a small complexity and hence computational complexity would be closer to the existing mechanism. 

The proposed DSDF method to estimate the discounted factor with existing QMIX method is explained in algorithm \ref{alg:complex}. The proposed DSDF has the following two advantages when compared with iterative penalization method. 

\begin{enumerate}
    \item DSDF method can dynamically change the discounted factor whereas the simpler method of penalizing it will only decrease it. Also, for the deterministic agents, the discounted factor is always $1$ and hence it may degrade those agents performance for accidental deviation. 
    
    \item DSDF method calculates the discounted factor of an agent relative to other agents. Hence it can consider complex interactions between agents while deciding discounted factor for the agents. 
\end{enumerate}

\begin{algorithm}
\caption{DSDF method to estimate discounted factor with QMIX}
\label{alg:complex}
\begin{algorithmic} 
\REQUIRE Initialize parameter vector $\theta_h$, hypernetwork parameters and $\theta$ (agents utility networks, maxing network, hyper network), Learning rate $\leftarrow$ $\alpha_\gamma$ and $\alpha_\theta$, $\mathcal{B}$ $\leftarrow$ $\{\}$
\REQUIRE step = 0, $\theta^- = \theta$
\WHILE{step < $\text{step}_{max}$}{}
\STATE $t = 0$, $s_0$ = Initial state
\WHILE{$t\neq$ terminal and $t<$ episode limit}{}
\FOR{each agent $i$}{}
\STATE $\tau_t^i=\tau_{t-1}^i \cup \{(o_t,u_{t-1}\}$  
\STATE $\epsilon = \text{epsilon-schedule (step)}$
\STATE  $u_a^t = \begin{cases}\begin{array}{cc}
    \underset{u_t^i}{\text{argmax}}\;\; Q(\tau_t^i,u_t^i) & \text{with probability}\; 1-\epsilon  \\
    \text{randint}(1,|U|) &  \text{with probability}\; \epsilon
\end{array}\end{cases}$
\ENDFOR
\STATE $s_{t+1} = p(s_{t+1}|s_t,\mathbf{u}_t)$
\STATE $\mathcal{B} = \mathcal{B} \cup \{(s_t,\mathbf{u}_t,r_t,s_{t+1}\}$
\STATE $t = t+1, step = step + 1$
\ENDWHILE
\IF{$|\mathcal{B}|>$ batch-size}{}
\STATE b $\leftarrow$ random batch of episodes from $\mathcal{B}$
\IF{$\theta_h$ not converged}{}
\STATE Update $\theta_h = \theta_h - \alpha_\gamma \triangledown_{\theta_h} (\Delta Q_{tot})^2$
\ENDIF
\STATE Update $\theta_\gamma = f_\gamma(O,\theta_h)$, where $O$ is the set of observations for all agents in the sampled batch.
\STATE Update $Q_{tot}$ using the latest updated $\theta_\gamma$. 
\STATE Update $\theta = \theta - \alpha_\theta \triangledown_{\theta} (\Delta Q_{tot})^2$
\ENDIF
\IF{update-interval steps have passed}{}
\STATE $\theta^- = \theta$
\ENDIF
\ENDWHILE
\end{algorithmic}
\end{algorithm}
\section{Results and Discussion}
\label{sec:results}
We performed experiments on the modified lbforaging \cite{paper:lbforaging} using the method discussed in Section \ref{sec:Neural} and results are presented in Section \ref{sec:res_env}. For the sake of stochasticity, we added a component which will decide whether to execute the action given by the policy or to take random action. 

\subsection{Environment}\label{sec:Env}
This environment contains agents and food resources randomly placed in a grid world. The agents navigate in the grid world and collect food resources by cooperating with other agents. We modified it by adding a couple of additions to the environment. 
\begin{enumerate}
    \item We added two additional actions '\textbf{Carry on}' and '\textbf{Leave}' to the agents. The '\textbf{Carry on}' action enables the agent to store the food resources which they can subsequently leave in another time step and/or state for consumption of another agent. The '\textbf{Leave}' action enables to agent to drop the resources which they consumed. 
    \item Each agent is given a target of the resources they need to consume. For this, we modify the reward function by adding targets to the them. Our eventual goal is to ensure all agents reach their targets. This means if some of the agents consumed in excess they must realize and give up the extra resources for benefit of other agents. 
\end{enumerate}
We placed $100$ food resources in the $30\times 30$ grid and we chosen $6$ agents to eat these resources. The episode is terminated either when there are no food resources available in the grid or number of time steps reached $500$. Here out of $6$ agents, three agents $3$ are deterministic (1,3,4) and $3$  are stochastic (2,5,6). The level of stochasticity $\beta$ means the will perform actions given by the policy with $1-\beta$ probability and will perform random actions with $\beta$ probability. The type of agents along with degree of stochasticity are shown in Table \ref{tab:types}
\begin{table}
    \centering
    \caption{Type of agents used in simulation}
    \begin{tabular}{|c|c|c|}
    \hline
    Agent Index & Type of Agent & Degree of stochasticity  \\ \hline
    1     &  Deterministic & NA \\ 
    2 & Stochastic & $0.2$ \\ 
    3 & Deterministic & NA \\ 
    4 & Deterministic & NA \\ 
    5 & Stochastic & $0.4$ \\ 
    6 & Stochastic & $0.6$ \\
    \hline
    \end{tabular}
    \label{tab:types}
\end{table}
We chosen the targets for the individual agents for two scenarios (i) When there are correct amount of food resources to achieve targets and (ii) When there are not enough resources to achieve the targets. 

\subsection{Results on Environment}
\label{sec:res_env}
In this section, we performed experiments on the above two cases with both the proposed DSDF method, iterative penalization method, QMIX method and IQL method. For this we utilized pymarl library used in \cite{paper:pymarl}. The individual targets for respective agents are $20,20,30,60,30$ and $60$. For both cases, we restrict sum of all food levels available in the grid to a number $T$ such that total sum of targets for individual agents is equal to $T$ in Case 1 and $>T$ in Case 2. \textbf{An important point to be noted that the value of $T$ is not known to any agent during training or in execution.} 

\textbf{Case 1 Enough Resources}: Here we chosen sum of targets to be $T=220$. Now, the agents have to collaborate within themselves to reach their respective targets.

\begin{figure}
\centering
\captionsetup{justification=centering}
\subfloat[Discounted factor prediction]{\includegraphics[width=4.5cm]{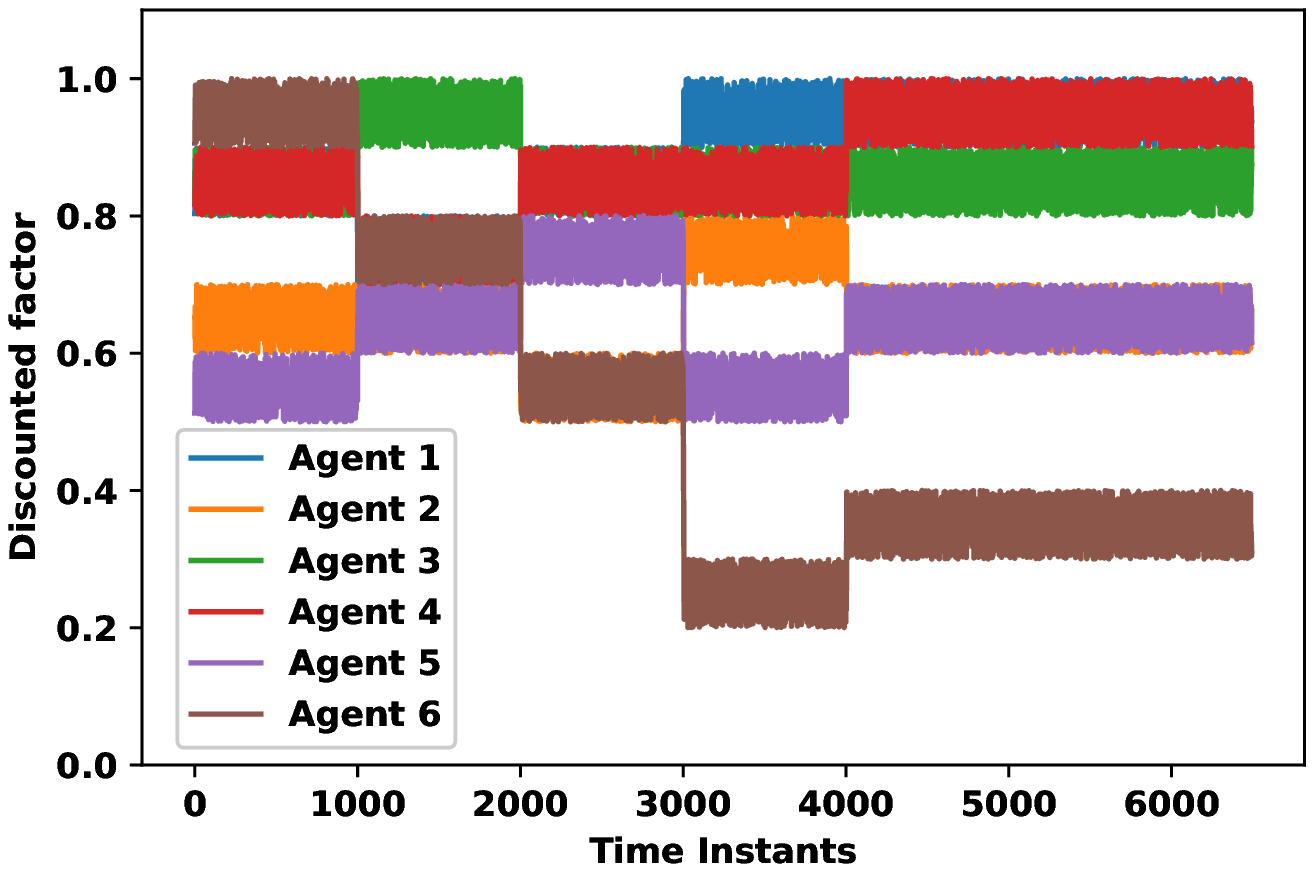}\label{fig:discounted_correct}}\hfil
\subfloat[Normalized global returns for correct resources]{\includegraphics[width=4.5cm]{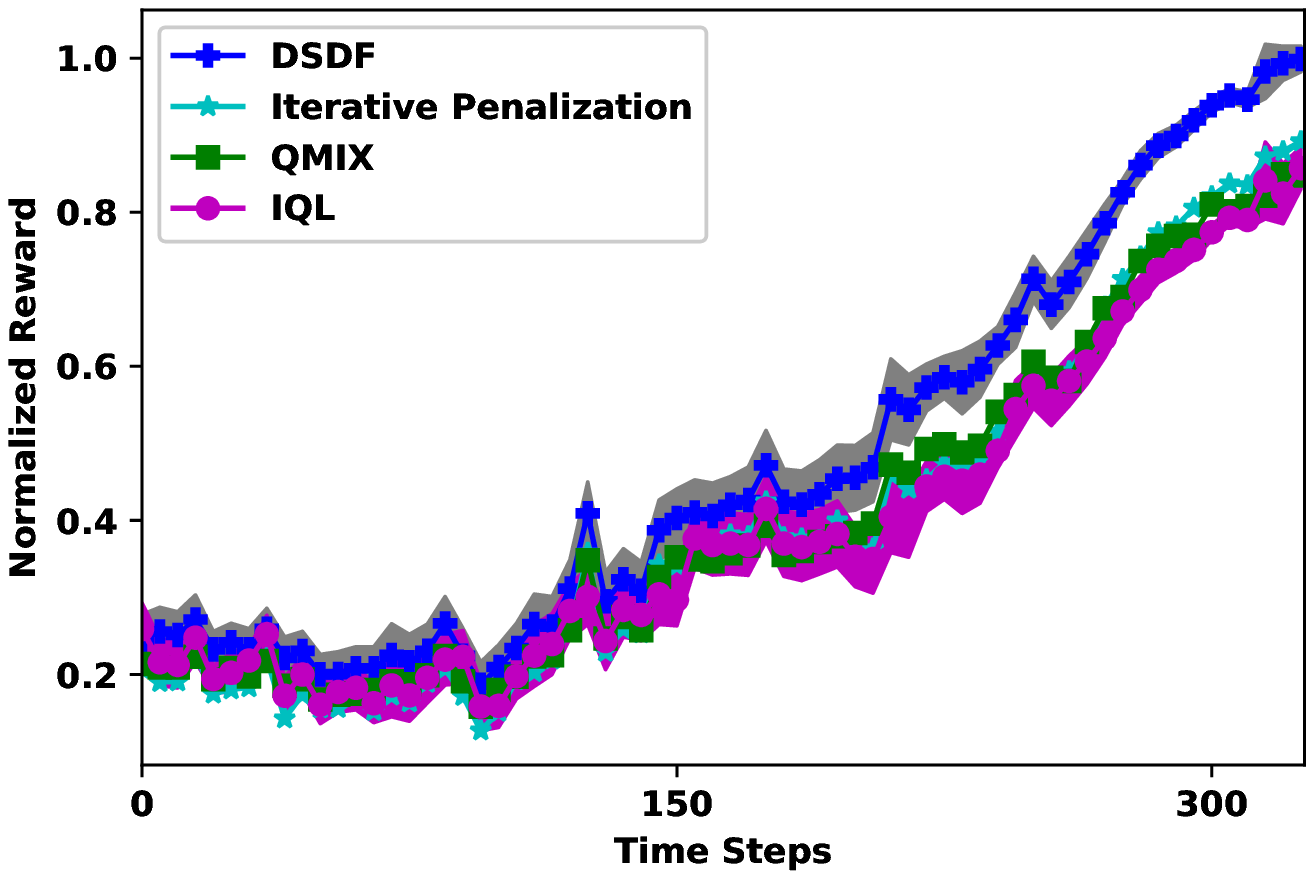}\label{fig:time_correct}}\hfil
\subfloat[Normalized global returns for scarce resources]{\includegraphics[width=4.5cm]{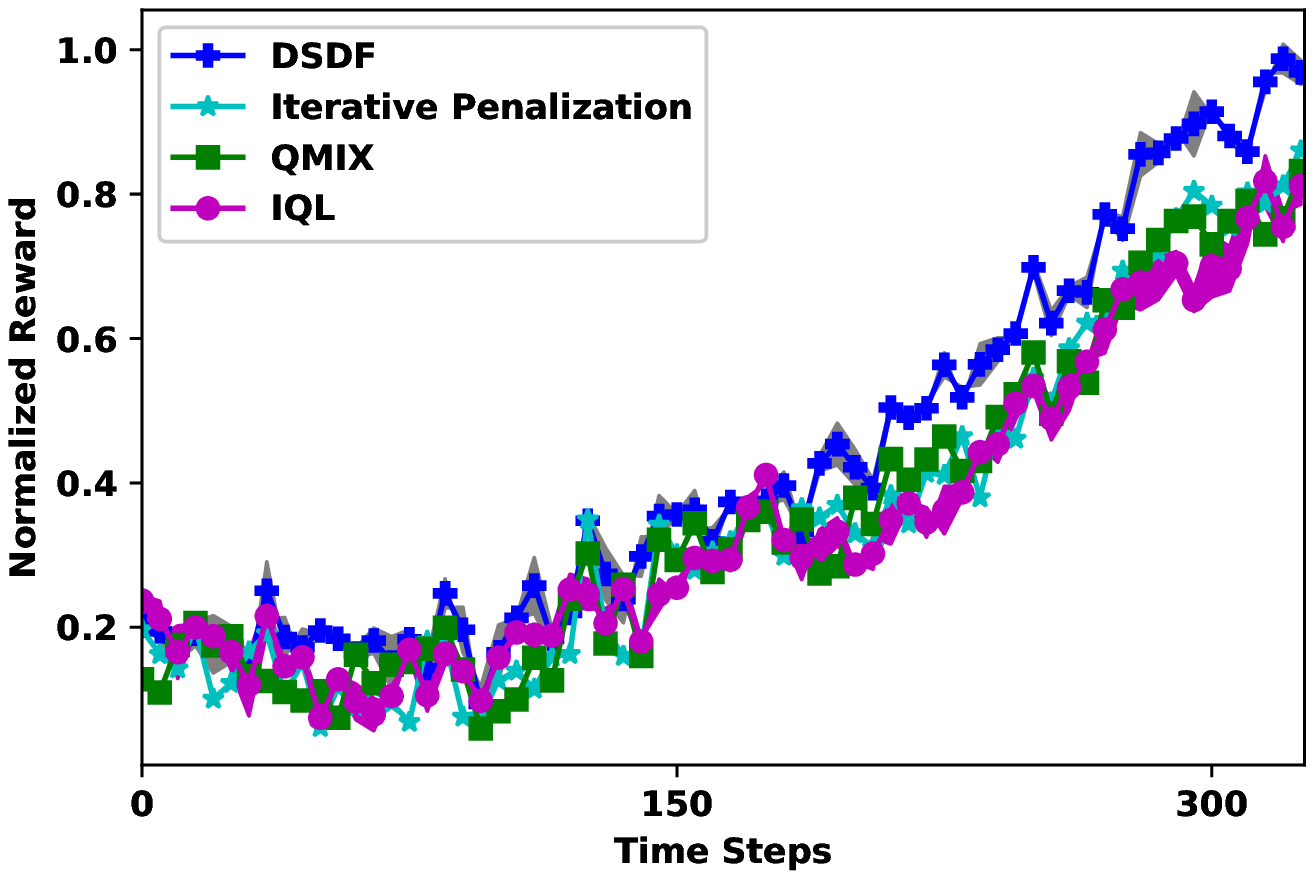}\label{fig:time_scarce}}
\caption{}
\end{figure}

The predictions of the discounted factor using DSDF method for all the time steps is shown in the Figure \ref{fig:discounted_correct}. As you can see the agents discounted factor changes whenever we update the discounted factor network. From the plot, we can observe the values got almost saturated after some update instants which shows our hypernetwork is converging and hence after this time we don't want to update the discounted factor hyper network. 

Also, one can observe the discounted factor values for deterministic agents are higher ($>= 0.9$) which suggests we can make the respective utility functions depend more on the future values. This is quite evident as these agents have to look more into future and decide current actions. On the other hand stochastic agents should choose lower discounted factor so that they won't take future values into consideration. From Figure \ref{fig:discounted_correct}, one can observe that the agents with higher degree of stochasticity i.e. they have less probability of executing actions given by the policy should have lesser discounted factor. This is agreeing with our assumption that the higher stochastic agents require lower discounted factor and vice-versa. 

\begin{figure}
\centering
\subfloat[Agent 1]{\includegraphics[width=4cm]{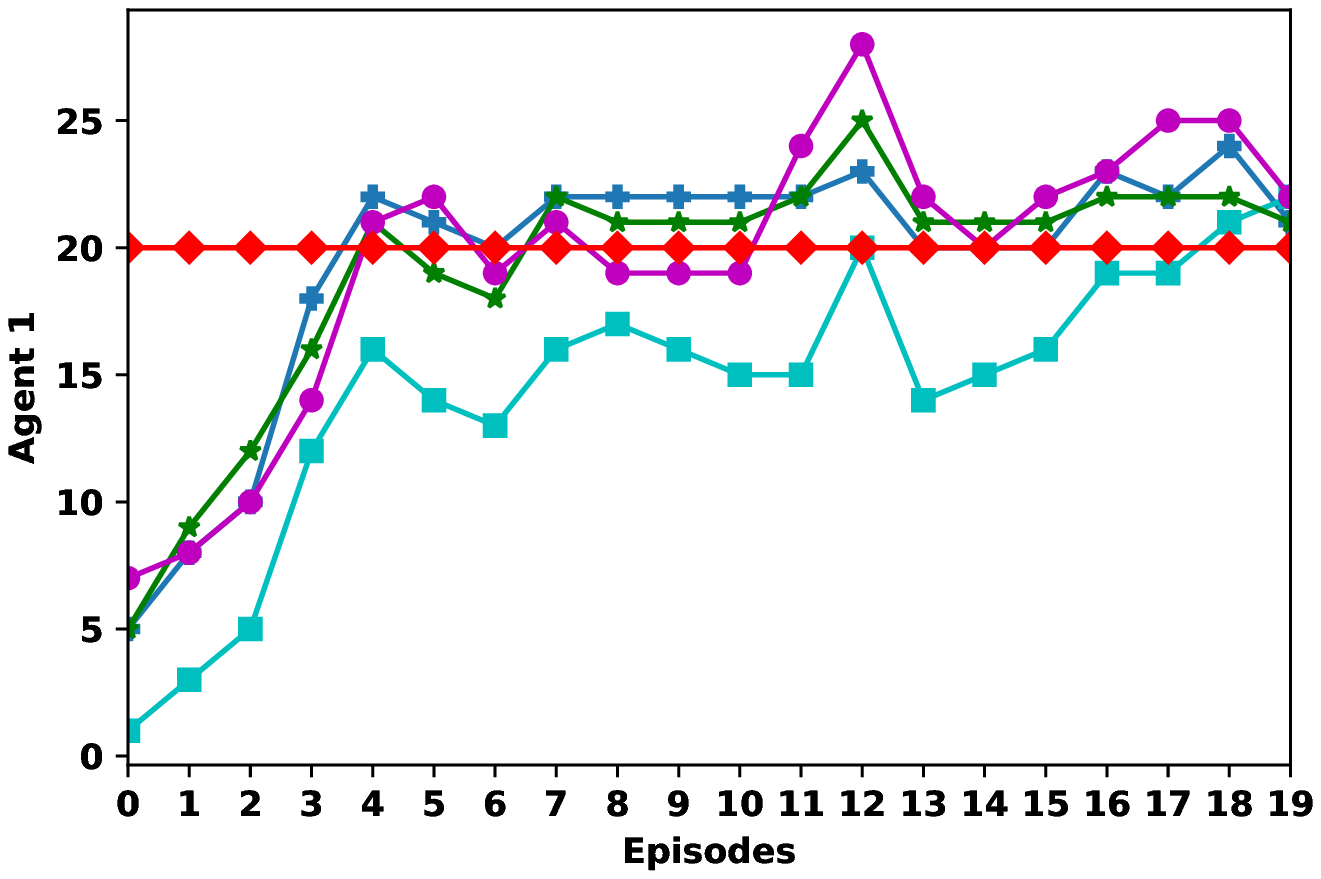}}\hfil
\subfloat[Agent 2]{\includegraphics[width=4cm]{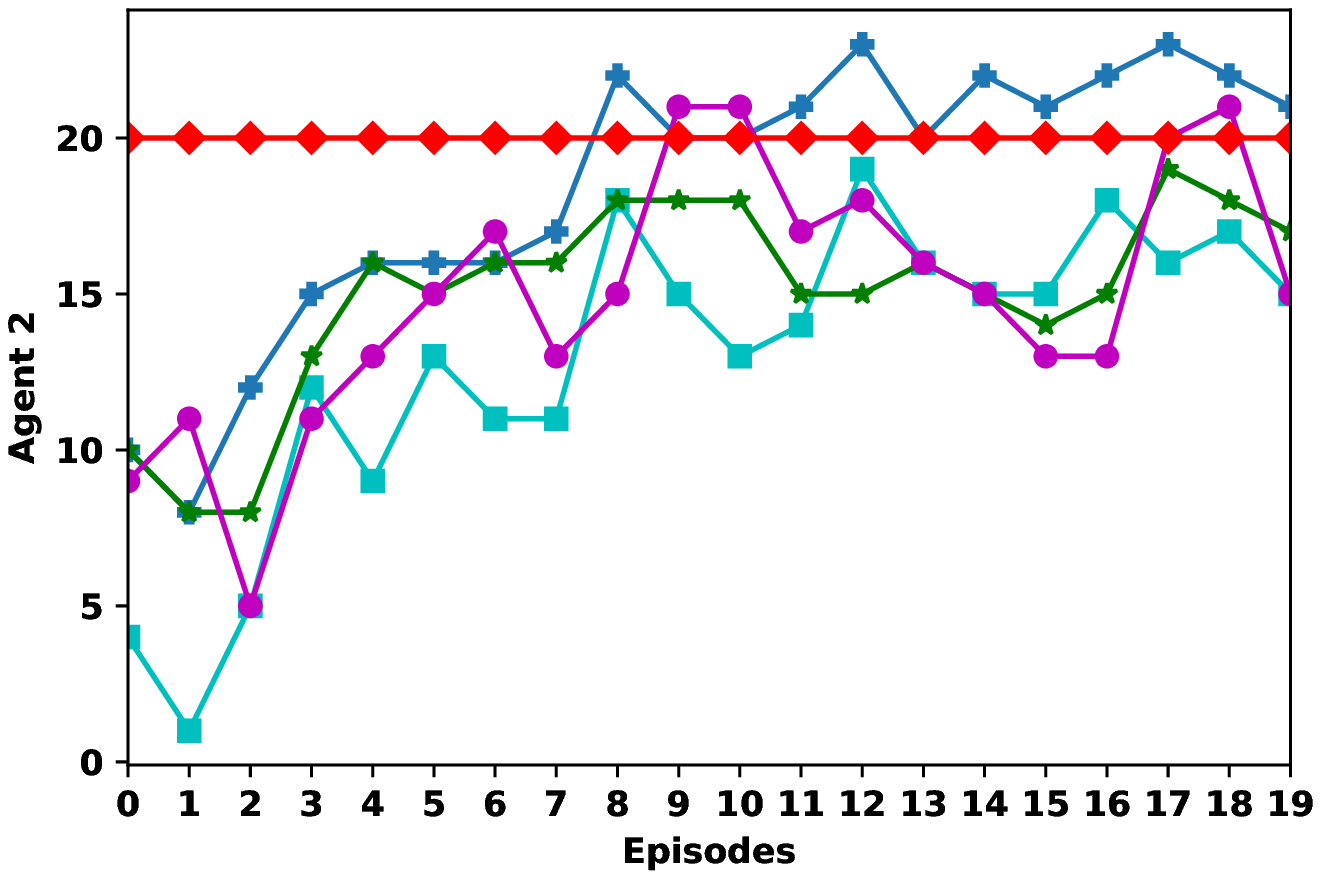}}\hfil 
\subfloat[Agent 3]{\includegraphics[width=4cm]{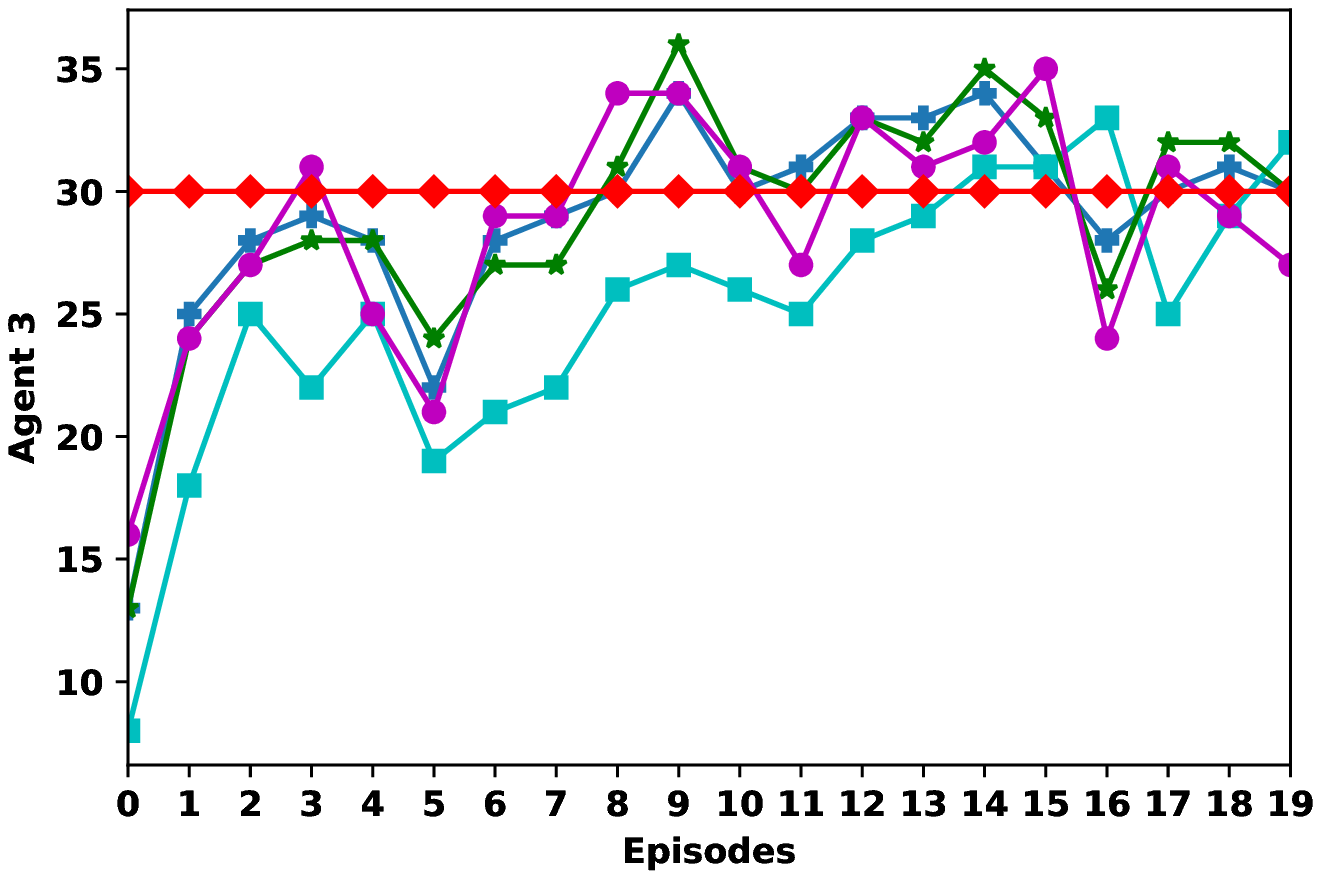}} 

\subfloat[Agent 4]{\includegraphics[width=4cm]{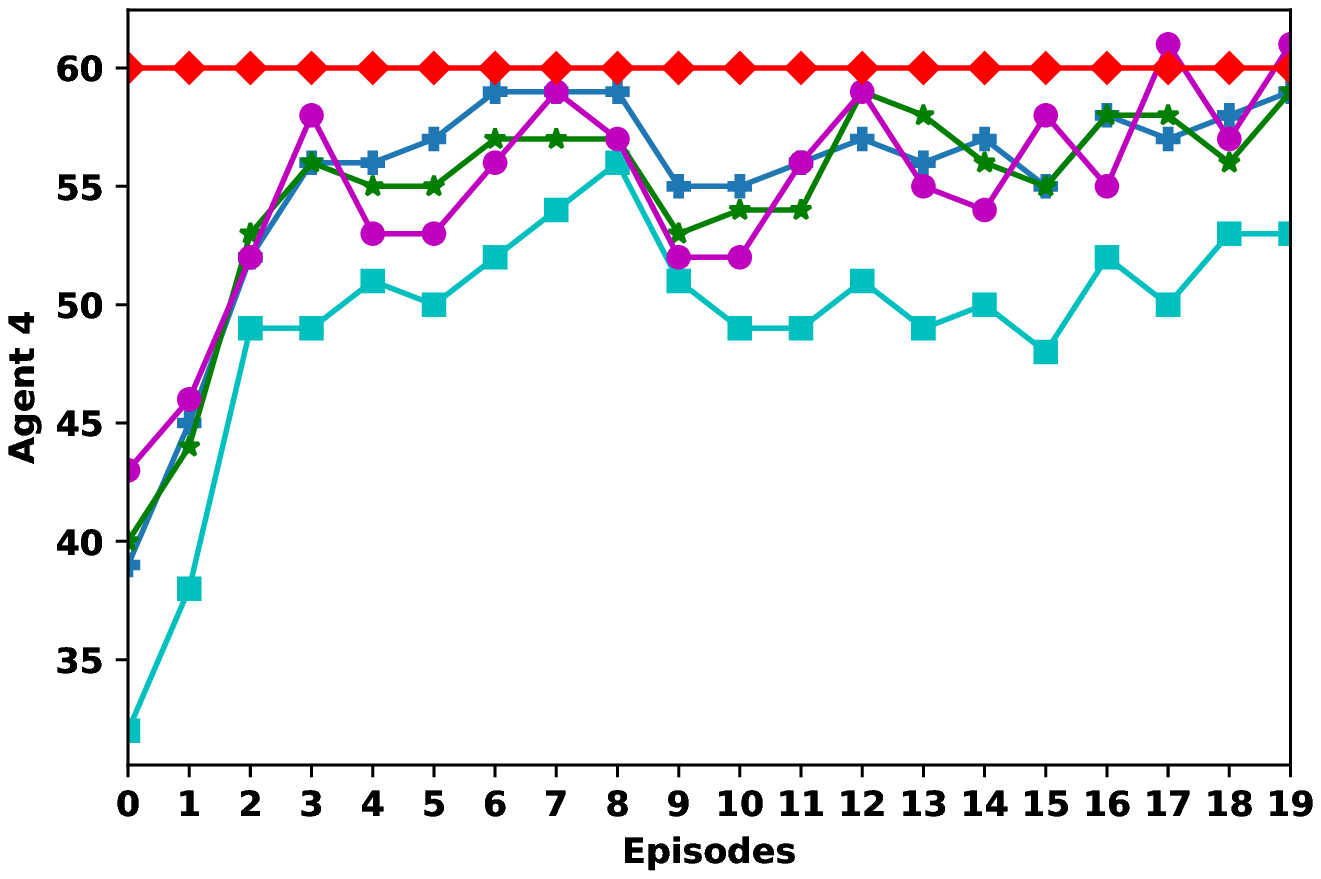}}\hfil   
\subfloat[Agent 5]{\includegraphics[width=4cm]{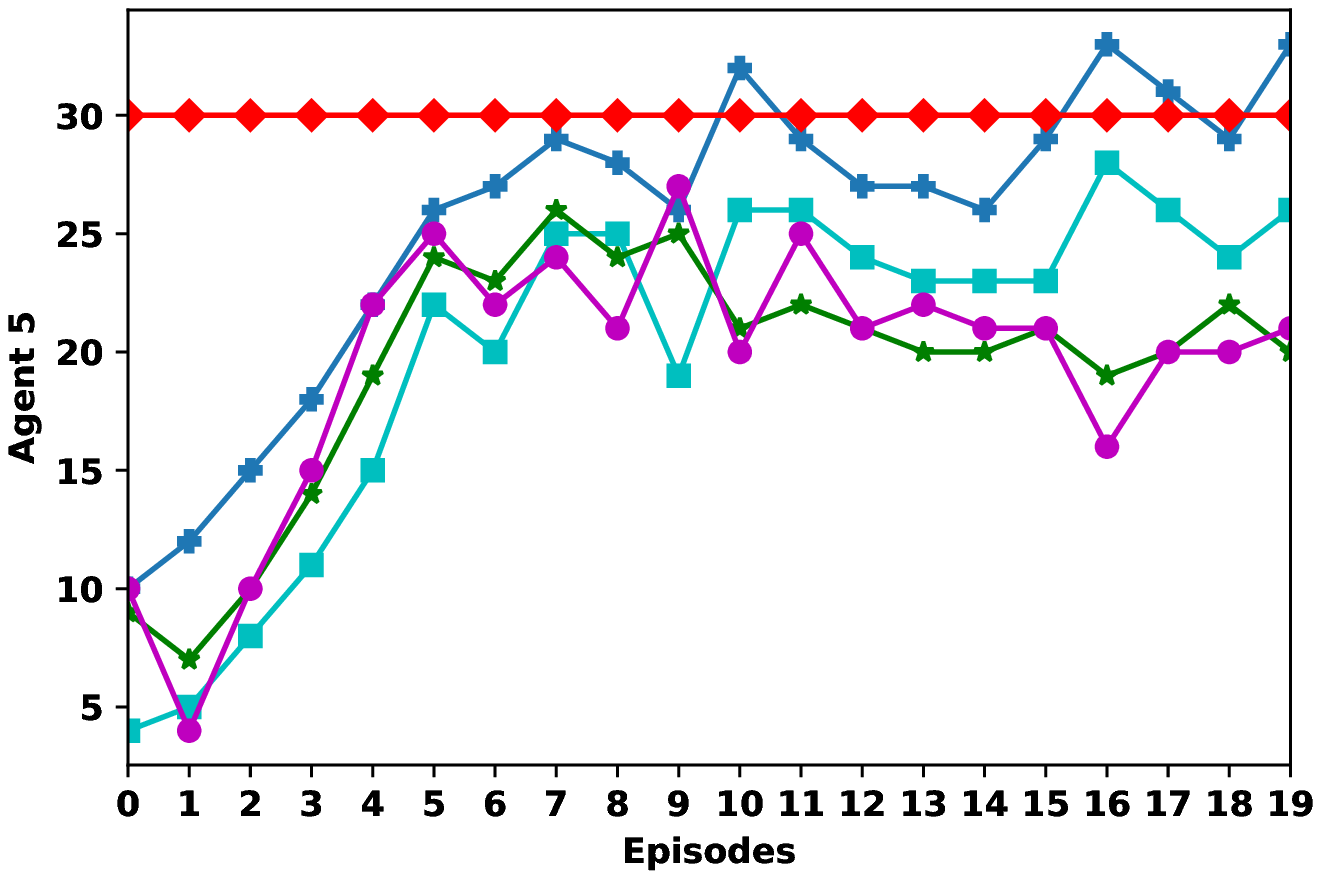}}\hfil
\subfloat[Agent 6]{\includegraphics[width=4cm]{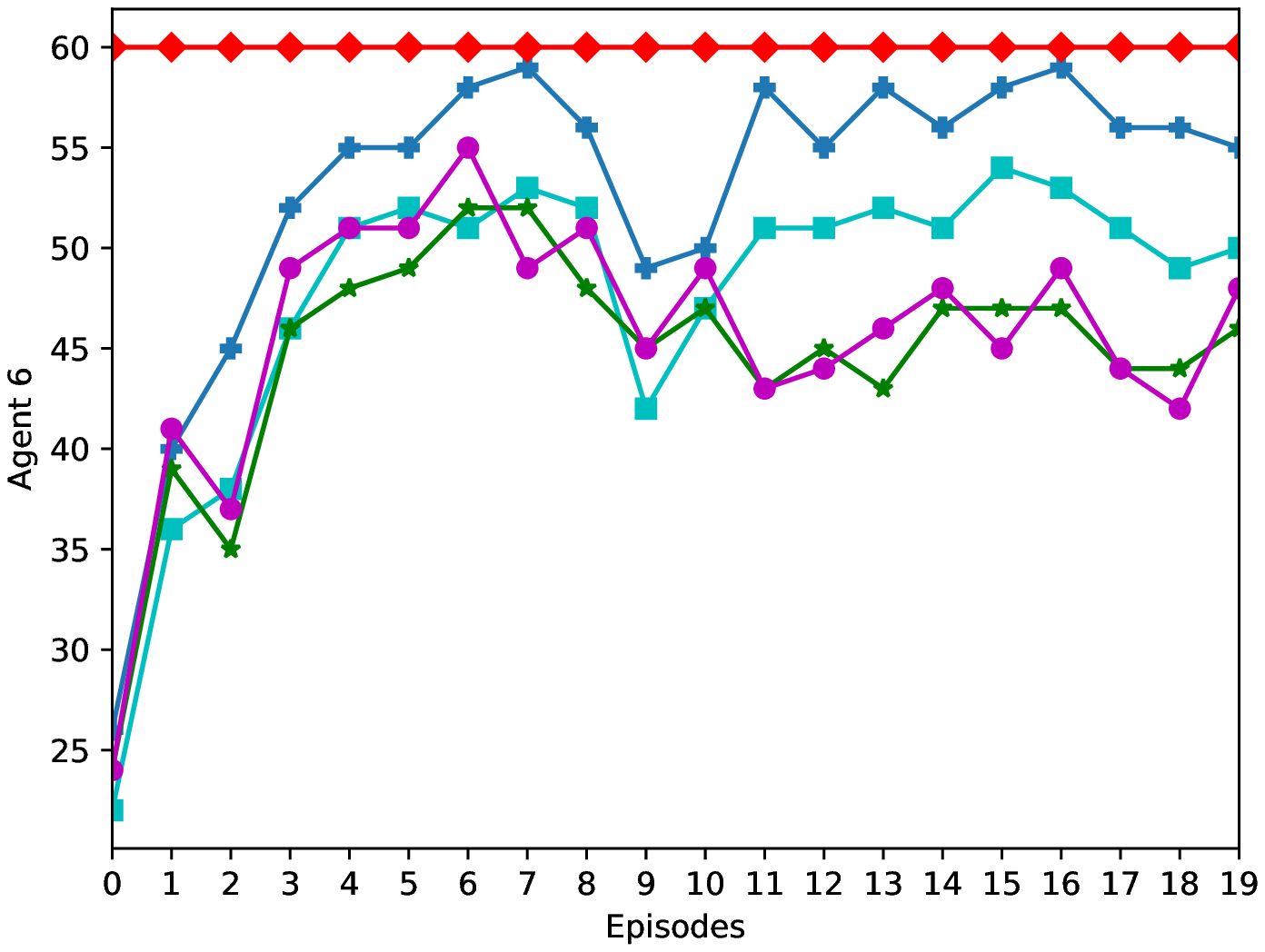}}

\subfloat{\includegraphics[scale=0.5]{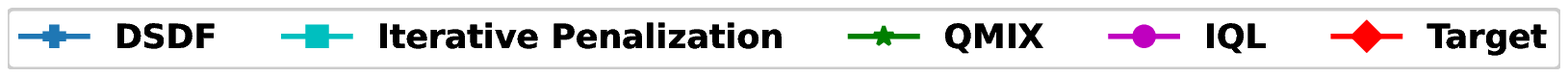}}\nonumber
\caption{Comparison of different methods for the case of correct resources}\label{figure:equal}
\end{figure}

Regarding the agent's performance, we gave the agents performance in last $10$ training episodes and $10$ execution episodes and results for individual agents are shown in Figure \ref{figure:equal}. In all the plots, first $10$ indices correspond to latest $10$ training episodes and next indices correspond to $10$ execution episodes. From the plots, it can be concluded that deterministic agents except agent $4$ trained using the proposed method reached their targets in all the execution episodes. On the other hand, the deterministic agents trained with vanilla QMIX and IQL have also reached their respective targets. The performance of the stochastic agents trained using the proposed DSDF method as well as iterative penalization method is well better than the vanilla QMIX and IQL. All the stochastic agents with smaller targets reached their targets while the agent with higher target settled closer to the target value.  For the case of deterministic agents, the agents trained using proposed DSDF method either performed well or shown comparable performance when compared with IQL and QMIX. On the other hand, the agents trained with iterative penalization method shown equal performance with QMIX and IQl. This is due to the fact that value of discounted factor is $1$ for deterministic agents, which means the agents depend more on the future and because of this it may spoil the agents performance. 

The mean reward obtained for every time step with 95\% confidence interval during evaluation is shown in Figure \ref{fig:time_correct}. From the plot, it is evident that the proposed DSDF method resulted in higher average reward when compared to iterative penalization method and other methods. The agents trained with iterative penalization method also fared comparably well when compared with QMIX and IQL methods.

\textbf{Case 2: Scarce Resources}: In this case, the sum of targets $T$ is chosen to be $200$ i.e. we do not have enough resources to satisfy all the agents.  In this case, we are skipping the discounted factor plot as it resembles to the one obtained in Figure \ref{fig:discounted_correct}.  

\begin{figure}
\centering
\subfloat[Agent 1]{\includegraphics[width=4cm]{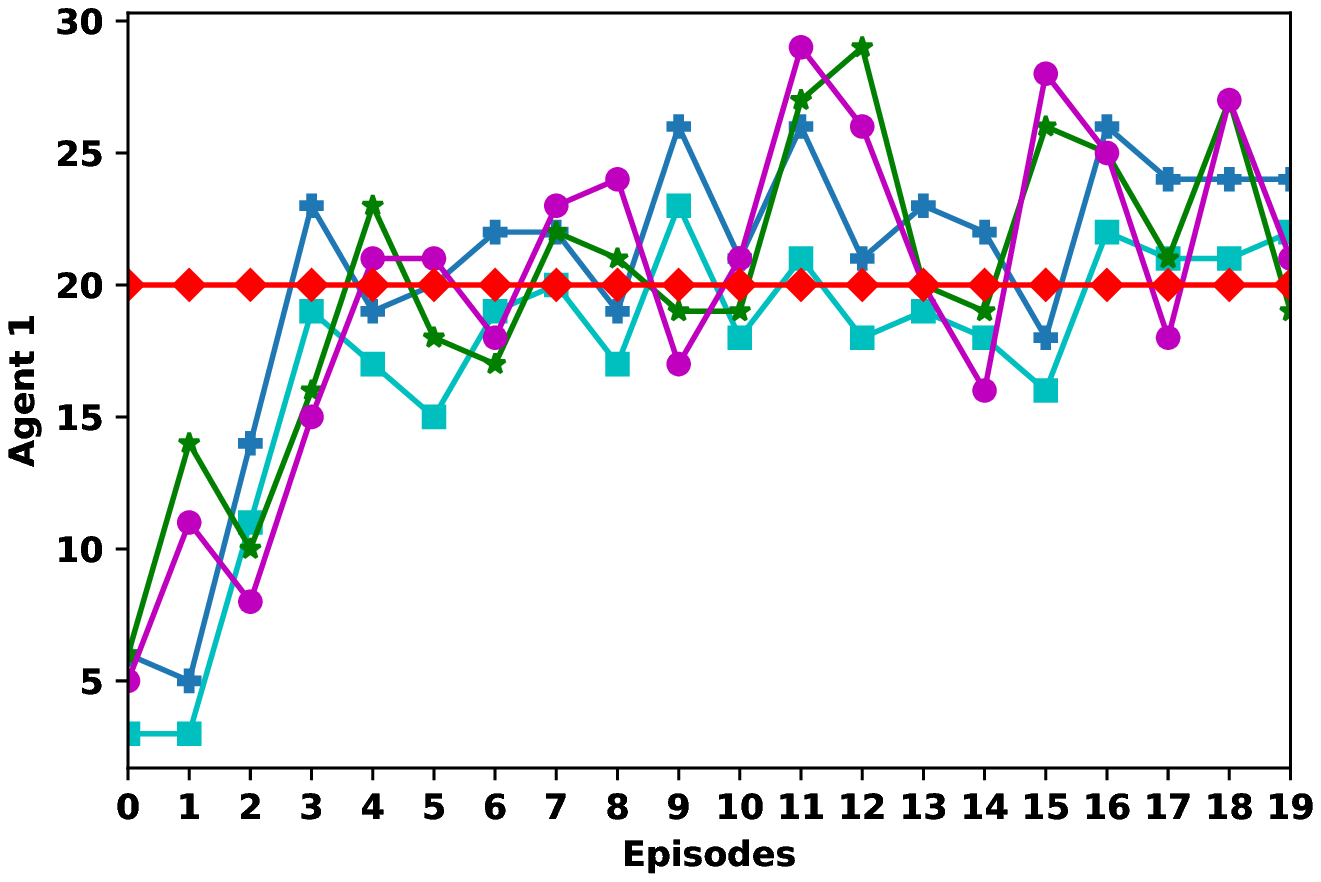}}\hfil
\subfloat[Agent 2]{\includegraphics[width=4cm]{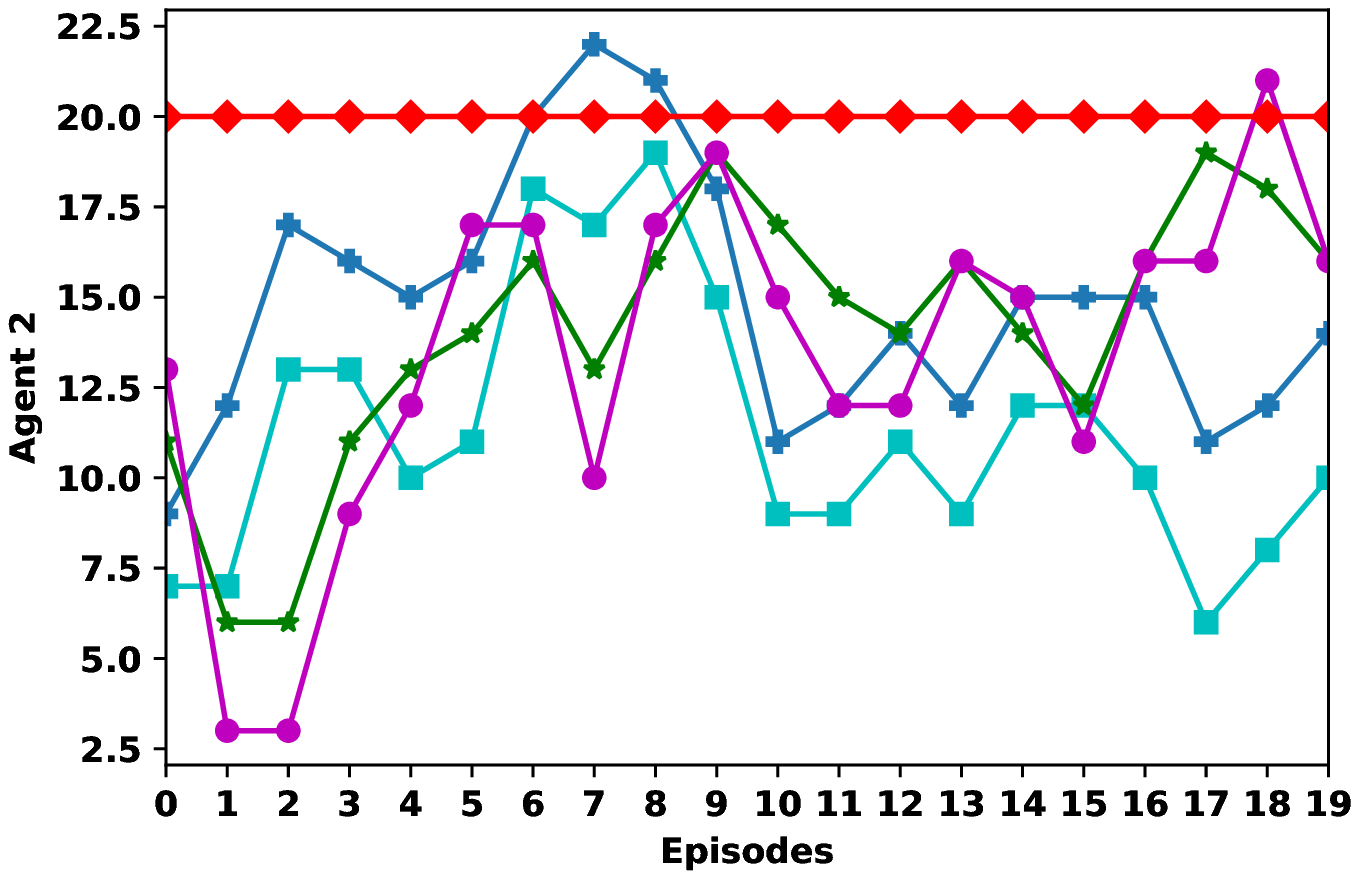}}\hfil 
\subfloat[Agent 3]{\includegraphics[width=3.75cm]{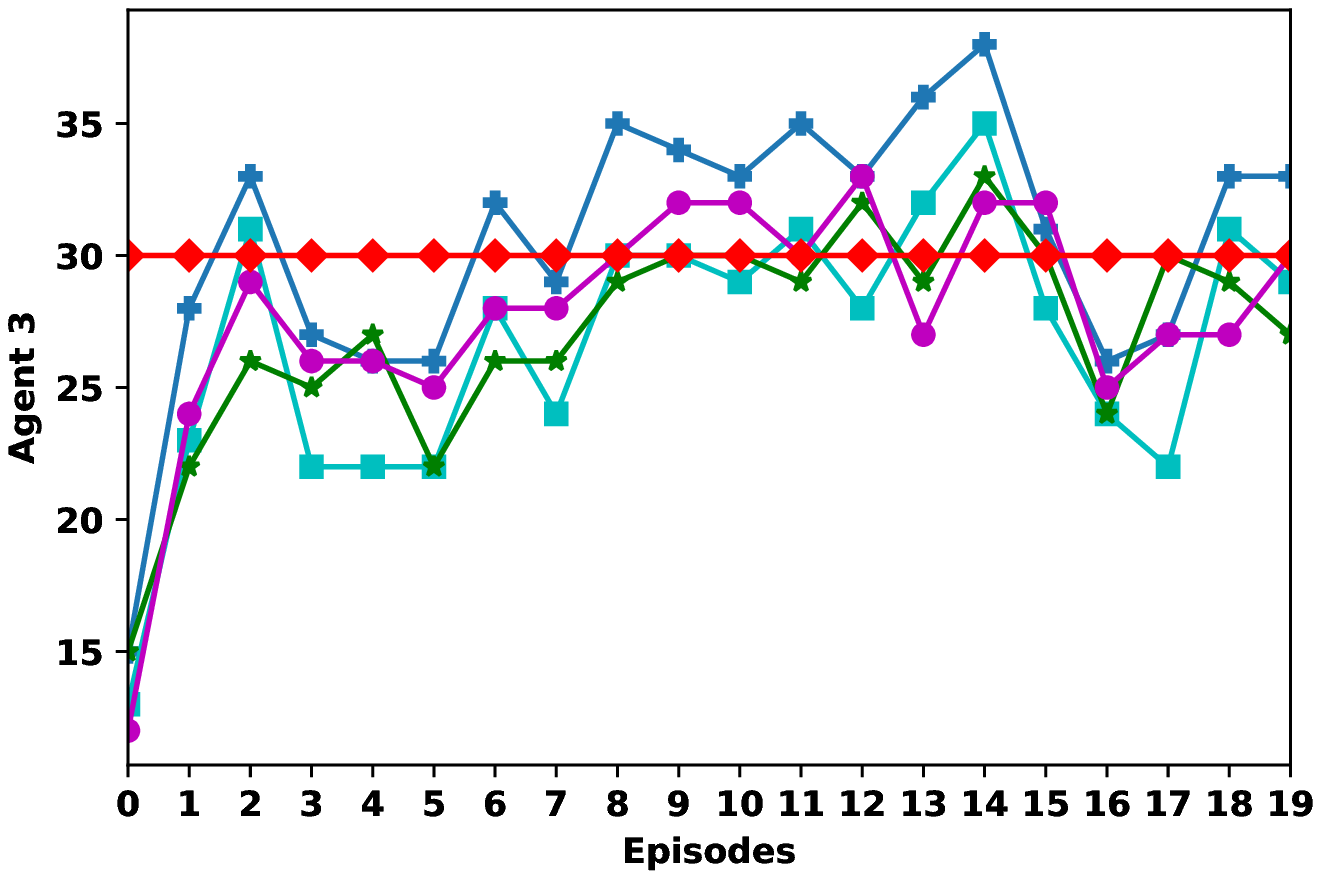}} 

\subfloat[Agent 4]{\includegraphics[width=4cm]{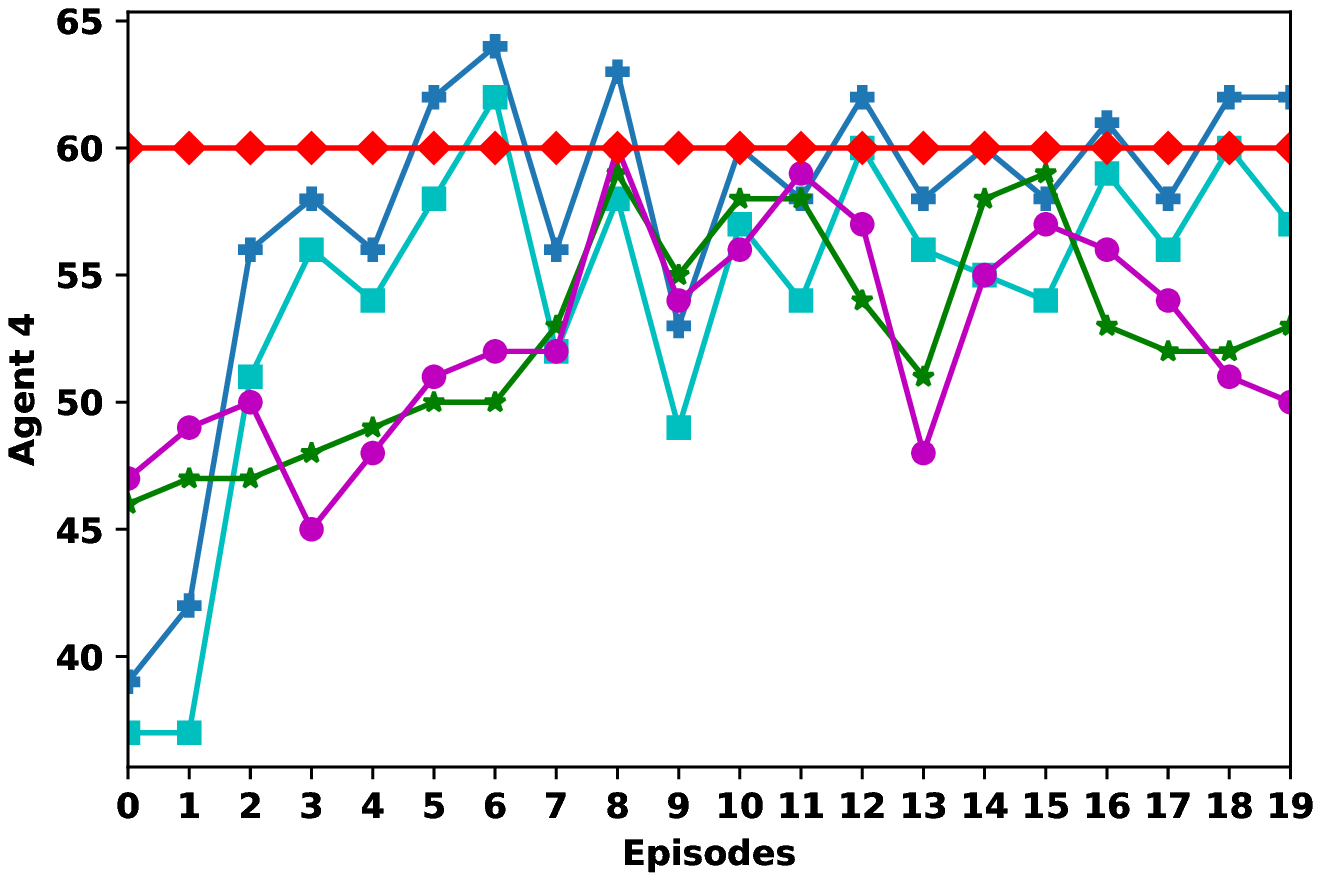}}\hfil   
\subfloat[Agent 5]{\includegraphics[width=4cm]{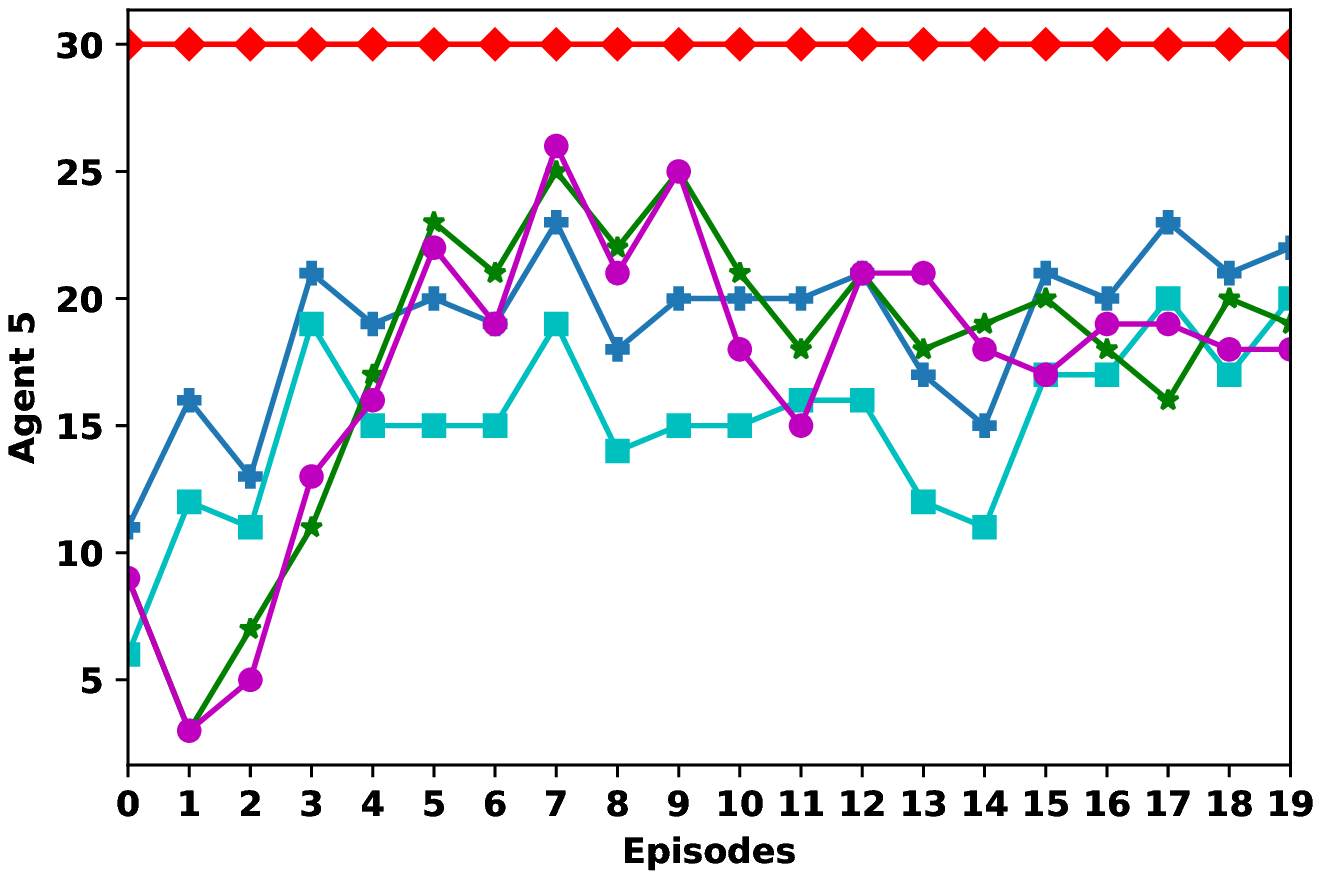}}\hfil
\subfloat[Agent 6]{\includegraphics[width=4cm]{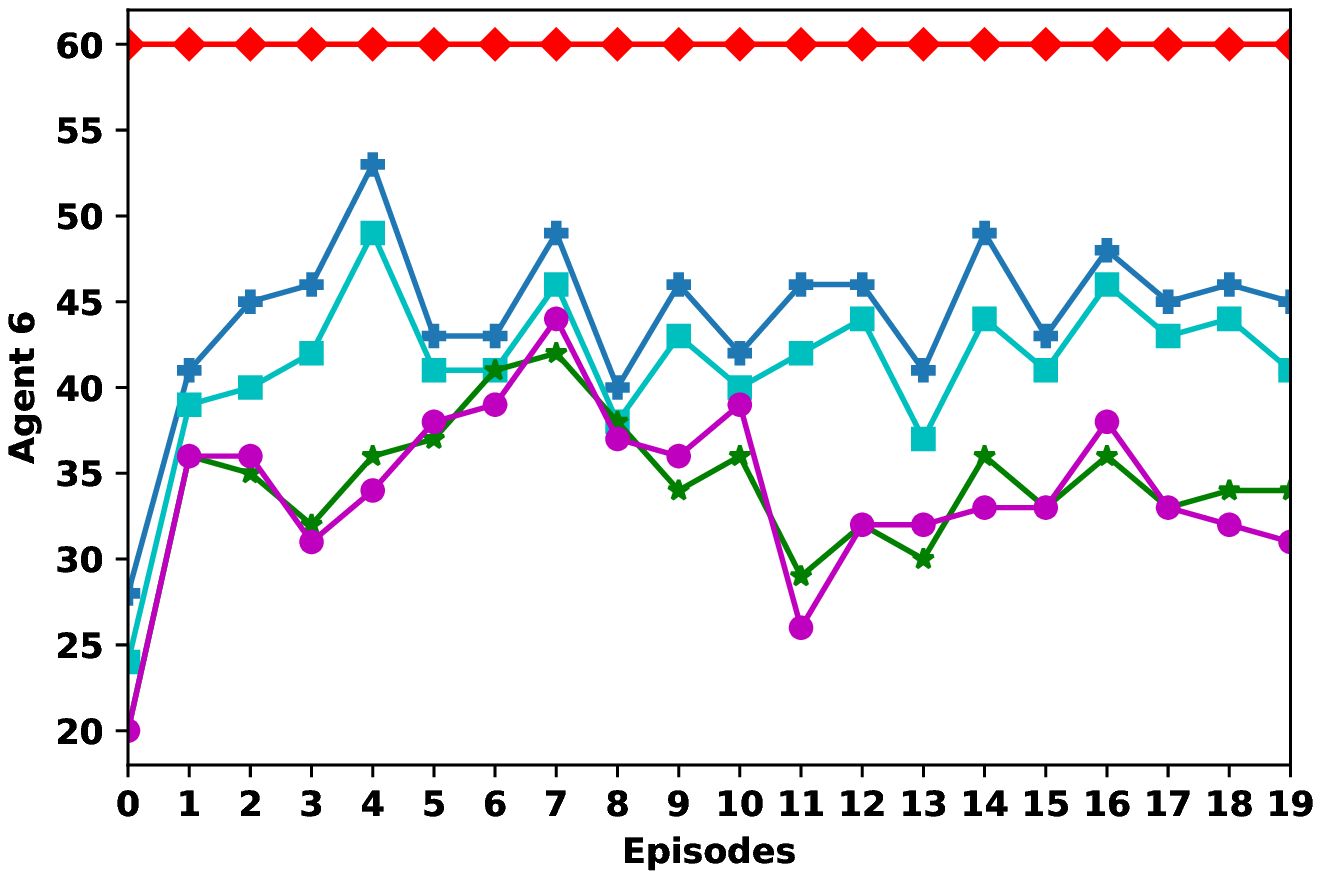}}

\subfloat{\includegraphics[scale=0.5]{legend.eps}}
\caption{Comparison of different methods for the case of scarce resources}\label{figure:scarce}
\end{figure}

Since we do not have enough resources the deterministic agents have to achieve all the targets and have to go beyond their targets since the dependency of the global reward on deterministic agents have to be made higher than stochastic agents. The agents performance for both the training and execution episodes are shown in Figure \ref{figure:scarce}. 
From the plots, one can observe all the deterministic agents trained with DSDF method reached their target and settled beyond their targets in the execution episodes. On other hand, agents trained with QMIX and IQL methods settled below the proposed method showing the efficacy of the proposed method. The proposed iterative penalization method also gave good results for stochastic agents with higher degree of stochasticity. 

Figure \ref{fig:time_scarce} shows the mean reward obtained with 95\% confidence interval in this scarce resources case with the agents trained with different methods. From the plot it is evident that the DSDF method resulted in good mean reward when compared with existing methods and also the iterative penalization method. In addition, the agents trained with iterative penalization method also fare comparably well when compared with existing methods. 

From the both scenarios, it can be concluded that the DSDF method results in better performance than the agents trained with QMIX and IQL method even when they are stochastic agents.  The proposed iterative penalization method also gave good results for stochastic agents with higher degree of stochasticity. For the case of deterministic agents, the performance of the iterative penalization is greater than or equal to the existing methods. 

\vspace{-0.5cm}
\section{Conclusion}\label{sec:conc}
In this paper, we propose a novel method DSDF to handle a mix of deterministic and stochastic agents which together learns a collaborative joint policy. Along with the above, an Iterative Penalization method is proposed, which though has lesser gains than DSDF. Our proposed methods can be combined with any state of art MARL algorithms without much impact to existing computation complexity.   

The proposed method is tested on the lbforaging environment in two different scenarios and demonstrated clear improvement in results when compared with QMIX and IQL methods both for individual and global returns. Future directions include extending the technique to situations where environment might also be stochastic in regions ( like oil patches on floor). To our knowledge, this is the first time, such an complicated interaction involving stochastic and deterministic agents have been explored. 

\bibliographystyle{plain}
\bibliography{References}

\end{document}